\documentclass[letterpaper]{article} 
\usepackage[preprint]{aaai2027}  
\usepackage[hyphens]{url}  
\usepackage{graphicx} 
\usepackage{natbib}  
\usepackage{caption} 
\usepackage{amsmath}
\usepackage{amssymb}
\usepackage{bm}
\usepackage{booktabs}
\usepackage{multirow}

\newcommand{\method}{CVT-GS}
\newcommand{\cL}{\mathcal{L}}

\title{\method{}: Learning to Simplify 3D Gaussian Splatting with Centroidal Voronoi Tessellation}
\author{
    Bingxian Li\textsuperscript{\rm 1},
    Yilong Li\textsuperscript{\rm 2},
    Jingliang Peng\textsuperscript{\rm 2},
    Peng-Shuai Wang\textsuperscript{\rm 2},
    Fei Zhu\textsuperscript{\rm 2}\corresponding,\\
    Guozheng Li\textsuperscript{\rm 1},
    Chi Harold Liu\textsuperscript{\rm 1},
    Guoping Wang\textsuperscript{\rm 2},
    Bo Pang\textsuperscript{\rm 2}\corresponding
}
\affiliations{
    \textsuperscript{\rm 1}Beijing Institute of Technology, Beijing, China\\
    \textsuperscript{\rm 2}Peking University, Beijing, China
}

\begin{document}
\maketitle
\flushbottom
\begin{abstract}
While 3D Gaussian Splatting (3DGS) has emerged as a powerful representation for real-time novel view synthesis, rendering high-fidelity scenes often relies on a massive number of Gaussian primitives, incurring substantial storage and computational overhead. 
Existing simplification techniques are largely intrusive, requiring training-time pruning, architectural modifications, or computationally expensive per-scene fine-tuning. 
These drawbacks limit their deployment on off-the-shelf pretrained models. 
In this paper, we propose \method{}, a novel optimization-free post-hoc simplification framework that directly compresses trained 3DGS scenes without sacrificing visual fidelity. 
Our approach first constructs spatially coherent cells over Gaussian centers via a geometry-aware Centroidal Voronoi Tessellation (CVT). Subsequently, a lightweight neural cell merger predicts the geometry and appearance of a single, highly representative Gaussian primitive for each cell under differentiable rendering supervision. By formulating simplification as a rendering-aware many-to-one merging process rather than naive primitive pruning, \method{} outputs a standard 3DGS scene that is seamlessly compatible with existing renderers.
Experiments on various datasets demonstrate the superiority of our method. Notably, when achieving a $100\times$ reduction in Gaussian points, our method operates $12\times$ faster than state-of-the-art methods while improving the PSNR by 1.3 dB.
\end{abstract}

\section{Introduction}
\label{sec:intro}

3D Gaussian Splatting (3DGS) has become a leading representation for 
novel view synthesis by combining explicit anisotropic Gaussian primitives with
efficient differentiable rasterization~\citep{kerbl20233dgs}. 
Compared with volumetric radiance fields~\citep{mildenhall2020nerf,barron2021mipnerf},
3DGS offers high visual fidelity, fast optimization, and efficient rendering,
but these advantages come with a structural cost. A high-quality scene often
requires hundreds of thousands to millions of Gaussian primitives, imposing
substantial storage, transmission, sorting, and rendering overhead and limiting
deployment on resource-constrained platforms.

To mitigate these limitations, recent 3DGS simplification and compression methods address this redundancy through pruning, structured representations, quantization, entropy coding, or neural attribute fitting~\citep{hanson2024pup3dgs,chen2025fastfeedforward,liu2025flexgs, chen2026pcgs,tang2026neuralgs,wang2025ghap}. These methods have made strong progress, but most of them are coupled with training-time optimization, representation redesign, specialized codecs, or scene-specific refinement. This leaves a practical gap because many standard 3DGS scenes are already trained, and users need a quick generic post-hoc method that \textit{preserves the original standard 3DGS format}. We then naturally ask, can we design a fast, generic post-hoc simplification framework that preserve original 3DGS format without extra optimization?

To answer this question, we must first formalize the post-hoc compression objective: reducing a pre-trained scene of $N$ Gaussians to a target count $M = \lceil \rho N \rceil$ under an aggressive compression ratio $\rho$, which could be as small as 0.001. 
The most intuitive baseline could be independent primitive pruning, or edge-collapse-like merging~\cite{xiong2026nanogs}. However, since $\rho$ could be extremely small, simply discarding primitives inevitably creates "holes" in scene coverage and fails to exploit the structural redundancy among spatially overlapping Gaussians. To preserve rendering fidelity at extreme compression rates, one must consolidate information rather than merely discard it.
Following this intuition, we shift the paradigm from naive pruning to rendering-aware many-to-one merging. Instead of deleting points, our formulation consolidates a spatial subset of input Gaussians into a newly predicted standard primitive, ensuring the simplified output remains seamlessly compatible with existing 3DGS renderers. However, such passway poses two central challenges. First, how to partition the scene into $M$ spatially coherent support regions, and second, how to convert each variable-sized support into a single Gaussian without sacrificing much visual fidelity.

We present \method{}, a post-hoc 3DGS simplification framework that separates many-to-one simplification into two steps, namely constructing mergeable support regions and predicting one output Gaussian for each support. For a prescribed output count, \method{} first partitions the input Gaussian centers into scene-level support regions using a geometry-aware Centroidal Voronoi Tessellation (CVT) inspired by classical geometry processing algorithms~\citep{du1999cvt,liu2009cvt,levy2010lpcvt}. 
Our CVT minimizes a global spatial distortion objective to allocate density-adaptive and spatially coherent support regions, bypassing the limitations of heuristic local neighborhoods~\citep{xiong2026nanogs} and tree-based grouping~\citep{wang2025ghap}. 
For each partitioned support, our lightweight neural cell merger, MergeNet, predicts a single representative Gaussian primitive in a single feed-forward pass. 
Because these merged primitives strictly preserve the standard 3DGS format, the resulting scenes remain plug-and-play with existing rasterizers. Experiments of multiple benchmark suggest our CVT-GS achievs a superior quality-speed balance under extreme compression ratio.

To conclude, our contributions include the following.
\begin{itemize}
    \item We introduce \method{}, a generic post-hoc framework that directly reduces the primitive count of pre-trained 3DGS scenes. This simplification process strictly preserves the standard representation and seamless renderer compatibility. 
    \item We propose a geometry-aware CVT formulation for constructing coherent support regions over Gaussian primitives, together with a lightweight neural cell merger that predicts one standard Gaussian for each support through a single feed-forward pass.
    \item Experiments on four 3DGS datasets demonstrate that \method{} achieves \textbf{higher PSNR} and better efficiency than prior state-of-the-art baselines, with up to \textbf{12.3$\times$} faster simplification under the same immediate-output protocol.
\end{itemize}

\section{Related Work}
\label{sec:related}

\subsection{Structured 3DGS Representations}
Structured 3DGS representations improve scene
organization, compactness, or deployment flexibility by changing how Gaussian
primitives are generated, parameterized, or decoded. 
Scaffold-GS predicts local
Gaussians from a structured scaffold representation~\citep{lu2024scaffold}, while compact or flexible
representations encode Gaussian attributes with neural fields, many-in-one
structures, or progressive bitstreams~\citep{liu2025flexgs,tang2026neuralgs,chen2026pcgs}. These methods broaden the
design space of 3DGS, but they typically introduce auxiliary structures or an
altered decoding process. In contrast, the proposed method starts from a pre-trained standard 3DGS scene and returns a reduced set of standard Gaussians, requiring no changes to the renderer or the original decoding pipeline.

\subsection{3DGS Simplification and Compression}
The large primitive count of 3DGS has also motivated pruning and coding
methods. Pruning-oriented methods use learned masks, constrained target counts,
uncertainty scores, or significance heuristics~\citep{lee2024compact3dgs,niedermayr2024compressed3dgs,fang2024minisplatting,
zhang2024lp3dgs,hanson2024pup3dgs,fan2024lightgaussian}, while coding-oriented
methods compress attributes through lightweight encodings, vector quantization,
context models, or entropy coding~\citep{girish2024eagles,navaneet2024compgs,chen2024hac,
chen2025fastfeedforward,dai2025decoupled,zhan2025cat3dgs}. Recent post-training
methods are closer to our setting. NanoGS performs training-free local pairwise
merging~\citep{xiong2026nanogs}, GHAP formulates Gaussian reduction from an
optimal-transport view~\citep{wang2025ghap}, and NeuralGS fits compact neural
fields but requires cluster-wise optimization and quality-restoring fine-tuning~\citep{tang2026neuralgs}. The proposed method instead constructs scene-level
CVT supports and predicts one standard Gaussian per cell with a shared neural
cell merger.

\subsection{CVT in Geometry Processing}
CVT is a classical tool in computational geometry for optimal
spatial partitioning and distribution-preserving sampling~\citep{du1999cvt}.
Unlike $k$-means or greedy clustering, CVT yields a principled
equilibrium in which each representative point optimally represents
its local region in a least-squares sense. It can also be extended
with anisotropic metrics and tensor-field guidance~\citep{liu2009cvt,levy2010lpcvt}.
Building on this geometry-processing perspective, we apply CVT to
trained 3DGS primitives. CVT determines which primitives form each
mergeable support, while a neural cell merger predicts one standard
Gaussian to represent each support.

\begin{figure*}[t]
\centering
\includegraphics[width=\textwidth]{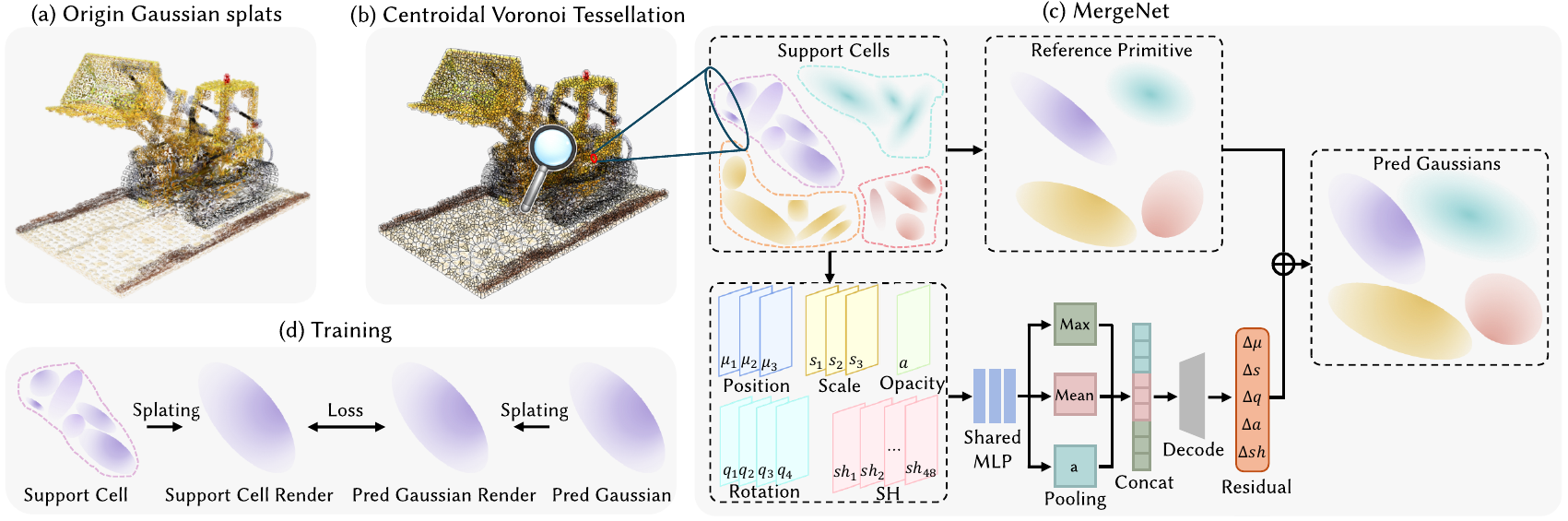}
\caption{Overview of our method. 
(a) The input is a trained 3DGS scene composed of original Gaussian splats. (b) Geometry-aware Centroidal Voronoi Tessellation (CVT) partitions Gaussian centers into $M=\lceil\rho N\rceil$ support cells.
(c) For each cell, an initial reference primitive is heuristically computed. MergeNet then predicts a residual update to refine this reference into a Gaussian point that better represent the overall appearance of cell.
(d) MergeNet is trained offline under local differentiable rendering supervision, aligning the rendered crops of the predicted Gaussian with those of the original cell. When applying our method, no parameters of the network will be updated.
}
\label{fig:pipeline}
\end{figure*}

\section{Method}
\label{sec:method}

\method{} addresses the post-hoc simplification of pre-trained 3DGS models. Given an input scene $S=\{G_i\}_{i=1}^{N}$ and a target simplification ratio $\rho$, we aim to output a standard 3DGS model $S^\star=\{G_k^\star\}_{k=1}^{M}$ with $M=\lceil\rho N\rceil$ primitives. Under this reduced primitive count, the objective is to preserve the rendering fidelity of the input model, defined as:
\begin{equation}
\min_{S^\star:\,|S^\star|=M}
\cL\!\left(\mathcal{R}(S^\star),\mathcal{R}(S)\right).
\label{eq:problem}
\end{equation}
where $\mathcal{R}$ denotes standard Gaussian splatting over evaluation views, and $\cL$ is an image-space rendering loss. 
The most naive approach might be try to directly optimizing Eq.~\eqref{eq:problem} for each input scene. However, such method would require scene-specific optimization, which is time-consuming and introduces substantial additional computation. 
Instead, \method{} decomposes the problem into scene-level support construction and feed-forward Gaussian prediction. Our geometry-aware CVT first converts the input Gaussian primitives into $M$ density-adaptive support cells in the 3D position space of Gaussian centers. Each cell is then treated as a variable-sized local set whose geometry and appearance are predicted by
MergeNet, a shared network trained once to output one standard Gaussian primitive per cell. Thus, the output primitives are newly predicted from grouped inputs rather than selected as a subset of $S$, and no per-scene fine-tuning is required.

In the following, we follow 3DGS~\cite{kerbl20233dgs} and define that each Gaussian point primitive $G_i$ contains position $\boldsymbol{\mu}_i$, covariance $\boldsymbol{\Sigma}_i$, scale $\mathbf{s}_i$, rotation $\mathbf{q}_i$, opacity $\alpha_i$, and appearance coefficients $\mathbf{f}_i$.

The following subsections discuss CVT support construction, cell feature construction, and MergeNet prediction with rendering supervision.

\subsection{Centroidal Voronoi Support Construction}
\label{sec:cvt_support}

Before predicting the output Gaussians, input primitives must be partitioned into support regions tailored for many-to-one merging. Fundamentally, this is a geometric allocation problem. If a support spans distant surfaces or disjoint structures, the merged primitive is forced to either unnaturally over-expand its covariance or sacrifice scene coverage. 
Conversely, relying on independent local neighborhoods fails to coordinate the $M$ outputs under a unified, scene-level objective.To overcome these dilemmas, we formulate support construction as a geometry-aware Centroidal Voronoi Tessellation (CVT) that allocates a fixed budget of supports across the pre-trained 3DGS distribution via a global variational objective. The CVT minimizes an opacity-weighted spatial energy, which quantifies the second-order spread that each output Gaussian must absorb. 
This minimization produces cells that are tightly bounded to their centroids while naturally adapting to the local density and opacity mass of the input Gaussians. Consequently, dense or high-opacity regions receive finer subdivisions, whereas sparser areas are efficiently represented by larger cells.

As geometric pre-processing, we follow NanoGS~\citep{xiong2026nanogs} and filter out Gaussians whose opacity is below $\tau_\alpha=0.1$. Please note that the prescribed output count remains unchanged, with $M=\lceil\rho N\rceil$ still computed from the original number of input primitives $N$. 
After that, CVT is computed in the 3D position space of Gaussian centers, with opacity used as sample mass. The resulting weighted samples define the empirical measure
\begin{equation}
\nu_{S}
=
\sum_{i\in\mathcal{I}}
\alpha_i\delta_{\boldsymbol{\mu}_i},
\label{eq:position_measure}
\end{equation}
where $\nu_S$ denotes this opacity-weighted empirical measure, $\alpha_i$ is the opacity of primitive $G_i$, and
$\delta_{\boldsymbol{\mu}_i}$ is the unit Dirac measure at
$\boldsymbol{\mu}_i$. In the integrals below,
$\mathbf{x}\in\mathbb{R}^3$ denotes a spatial location.

Given $M$ sites $C=\{\mathbf{c}_k\}_{k=1}^{M}$, where
$\mathbf{c}_k\in\mathbb{R}^3$ is the site of the $k$-th cell, the Voronoi region of that site is
\begin{equation}
\begin{aligned}
 \Omega_k(C)=
\big\{\mathbf{x}\in\mathbb{R}^3:
&\ \|\mathbf{x}-\mathbf{c}_k\|_2^2 \\
&\le
\|\mathbf{x}-\mathbf{c}_j\|_2^2,\ 
\forall j\in\{1,\ldots,M\}
\big\}.
\end{aligned}
\label{eq:voronoi_region}
\end{equation}
The support construction is formulated as optimal quantization of
$\nu_S$, with CVT energy $E_{\mathrm{CVT}}$ defined as
\begin{equation}
\begin{aligned}
E_{\mathrm{CVT}}(C)
&=
\sum_{k=1}^{M}
\int_{\Omega_k(C)}
\|\mathbf{x}-\mathbf{c}_k\|_2^2
\,d\nu_{S}(\mathbf{x}).
\end{aligned}
\label{eq:cvt_continuous}
\end{equation}
The energy in Eq.~\eqref{eq:cvt_continuous} couples all cells through one
scene-level distortion. As a result, the support of each output primitive is
determined relative to all other supports, rather than by an isolated local
neighborhood decision.

Since $\nu_{S}$ is supported on discrete Gaussian positions, the method optimizes the
cell memberships of remaining primitives. For fixed sites, the cell assignment
$a(i)\in\{1,\ldots,M\}$ of primitive $i$ and the member set $V_k$ of the $k$-th cell are given by the 3D Voronoi map
\begin{equation}
a(i)=\operatorname*{argmin}_{j\in\{1,\ldots,M\}}\|\boldsymbol{\mu}_i-\mathbf{c}_j\|_2^2,\quad
V_k=\{i\in\mathcal{I}\mid a(i)=k\}.
\label{eq:cvt_assignment}
\end{equation}
Thus, $V_k$ contains the indices of input Gaussians assigned to the $k$-th support cell after opacity filtering.
Substituting Eq.~\eqref{eq:position_measure} into Eq.~\eqref{eq:cvt_continuous} yields the discrete CVT objective
\begin{equation}
E_{\mathrm{CVT}}
=
\sum_{k=1}^{M}
\sum_{i\in V_k}
\alpha_i\|\boldsymbol{\mu}_i-\mathbf{c}_k\|_2^2 .
\label{eq:cvt_discrete}
\end{equation}
For fixed memberships $V_k$, minimizing Eq.~\eqref{eq:cvt_discrete} gives the closed-form Lloyd site update
\begin{equation}
\mathbf{c}_k
=
\frac{
\sum_{i\in V_k}
\alpha_i\boldsymbol{\mu}_i
}{
\sum_{i\in V_k}
\alpha_i
}.
\label{eq:cvt_update}
\end{equation}
Alternating Eq.~\eqref{eq:cvt_assignment} and Eq.~\eqref{eq:cvt_update} performs Lloyd relaxation~\citep{lloyd1982least} on $\nu_S$.

For Gaussian simplification, Eq.~\eqref{eq:cvt_discrete} has a direct geometric interpretation because it measures the weighted second-order spatial spread that each output primitive must absorb within its cell. Lowering this quantity yields tighter and more coherent supports, reducing the burden on the subsequent single-Gaussian prediction. The centroidal update in Eq.~\eqref{eq:cvt_update} further enforces that each site is the opacity-weighted center of its assigned primitives, which is precisely the density-adaptive behavior needed when a target number of output Gaussians must cover a highly non-uniform 3DGS distribution.
This weighted Lloyd relaxation is implemented with the Geogram library~\citep{geogram}. 
Let $N_c=|\mathcal{I}|$ be the number of remaining primitives. For $T$ Lloyd iterations, the assignment step has the conservative upper bound $\mathcal{O}(T N_c M)$; in practice, Geogram accelerates nearest-site queries with spatial search structures.

\subsection{Cell Feature Construction} \label{sec:cell_feature_const}
After CVT, $V_k$ denotes the index set of input Gaussians assigned to the $k$-th support cell after opacity filtering. The variable-sized member set $\{G_i\}_{i\in V_k}$ is used to predict one output primitive $G_k^\star$. 
Before applying MergeNet, the cell is normalized by a deterministic reference primitive $\bar{G}_k$. This reference provides a stable local coordinate system and initial parameter scale for residual prediction.

Let $\omega_i=\alpha_i/\sum_{j\in V_k}\alpha_j$ be the normalized opacity weight inside $V_k$. The reference position and appearance are weighted averages, $\bar{\boldsymbol{\mu}}_k=\sum_{i\in V_k}\omega_i\boldsymbol{\mu}_i$ and $\bar{\mathbf{f}}_k=\sum_{i\in V_k}\omega_i\mathbf{f}_i$, where $\mathbf{f}_i$ denotes the appearance coefficients of primitive $G_i$. 
The reference covariance is obtained from the weighted second spatial moment of the Gaussian mixture in the cell
\begin{equation}
\begin{aligned}
\bar{\boldsymbol{\Sigma}}_k=
\underbrace{\sum_{i\in V_k}\omega_i\boldsymbol{\Sigma}_i}_{\text{within-splat shape}}
&+
\underbrace{\sum_{i\in V_k}\omega_i
(\boldsymbol{\mu}_i-\bar{\boldsymbol{\mu}}_k)
(\boldsymbol{\mu}_i-\bar{\boldsymbol{\mu}}_k)^\top}_{\text{between-splat spread}} .
\end{aligned}
\label{eq:representative_covariance}
\end{equation}
The scale $\bar{\mathbf{s}}_k$ is obtained from the square roots of the sorted eigenvalues of $\bar{\boldsymbol{\Sigma}}_k$, and the rotation $\bar{\mathbf{q}}_k$ is obtained from the corresponding eigenvectors. Opacity is initialized by probabilistic composition, $\bar{\alpha}_k=1-\prod_{i\in V_k}(1-\alpha_i)$.

Using $\bar{G}_k=(\bar{\boldsymbol{\mu}}_k,\bar{\mathbf{s}}_k,\bar{\mathbf{q}}_k,\bar{\alpha}_k,\bar{\mathbf{f}}_k)$, each member primitive $G_i$ with $i\in V_k$ is described by a reference-relative descriptor
$\mathbf{z}_{ik}$. 
It contains the normalized local position $(\bar{\mathbf{R}}_k)^\top(\boldsymbol{\mu}_i-\bar{\boldsymbol{\mu}}_k) \oslash\bar{\mathbf{s}}_k$, relative log-scale
$\log\mathbf{s}_i-\log\bar{\mathbf{s}}_k$, relative rotation $\bar{\mathbf{q}}_k^{-1}\otimes\mathbf{q}_i$, opacity terms $(\alpha_i,\omega_i)$, and the appearance residual $\mathbf{f}_i-\bar{\mathbf{f}}_k$, where $\bar{\mathbf{R}}_k$ is the rotation
matrix induced by $\bar{\mathbf{q}}_k$ and $\oslash$ denotes element-wise division; $\otimes$ denotes quaternion multiplication. 
Quaternion signs are aligned to $\bar{\mathbf{q}}_k$ before relative rotations are formed. 
These reference-relative features provide a normalized local description of each cell, allowing the shared MergeNet to operate across scenes and simplification ratios.

\subsection{MergeNet Prediction and Loss} \label{sec:method_merge_net}
MergeNet is a permutation-invariant set predictor that maps each CVT cell to one
standard Gaussian primitive. Given the reference-relative descriptors
$\{\mathbf{z}_{ik}\}_{i\in V_k}$, a shared pointwise MLP $\phi$ embeds every cell
member. Mean pooling, max pooling, and opacity-weighted pooling aggregate the
variable-sized set into a fixed-dimensional cell code $\mathbf{h}_k$. A decoder
MLP $\psi$ then predicts a residual update $\Delta G_k$ with respect to the
reference primitive
\begin{equation}
\Delta G_k=\psi(\mathbf{h}_k),\qquad
G_k^\star=\bar{G}_k\oplus\Delta G_k .
\label{eq:residual_decode}
\end{equation}
Here $\oplus$ denotes reference-relative composition in the standard 3DGS
parameter space, where positions are predicted in the local frame of $\bar{G}_k$,
scales and opacities are updated in log-scale and logit-opacity domains,
rotations are composed by quaternion increments, and appearance coefficients are
updated additively. This parameterization keeps the prediction normalized across
cells of different spatial extents while preserving a valid standard 3DGS
primitive. The three pooling operators are complementary. Mean pooling captures
average cell statistics, max pooling preserves salient member responses, and
opacity-weighted pooling emphasizes primitives with larger visual contribution.

Training is performed with local differentiable rendering supervision. For each
CVT cell $V_k$, let $S_k=\{G_i\}_{i\in V_k}$ denote the input member set. For
a sampled local crop $\pi$ associated with the $k$-th cell, $S_k$ and its
one-primitive prediction $G_k^\star$ are rendered with the same rasterizer and
crop window, producing the reference crop $I_{k,\pi}$ and predicted crop
$\hat{I}_{k,\pi}$. In addition to the standard rendering loss, a gradient
consistency term encourages local edge and texture preservation
\begin{equation}
\mathcal{L}_{\mathrm{grad}}
=
\|\nabla_x\hat{I}_{k,\pi}-\nabla_x I_{k,\pi}\|_1
+
\|\nabla_y\hat{I}_{k,\pi}-\nabla_y I_{k,\pi}\|_1 .
\label{eq:grad_loss}
\end{equation}
\begin{equation}
\mathcal{L}_{\mathrm{train}}
=(1-\lambda)\mathcal{L}_{1}
+\lambda\mathcal{L}_{\mathrm{SSIM}}
+\lambda_g\mathcal{L}_{\mathrm{grad}} .
\label{eq:loss_total}
\end{equation}
Here $\nabla_x$ and $\nabla_y$ denote finite-difference image gradients.
$\mathcal{L}_{1}$ is the mean absolute pixel error,
$\mathcal{L}_{\mathrm{SSIM}}=1-\mathrm{SSIM}$, and $(\lambda,\lambda_g)$ are
loss weights. Losses are averaged over sampled cells and crops, with the
original cell rendering used as a fixed target; only MergeNet parameters are
updated.At deployment, CVT-GS exports the simplified scene as a standard 3DGS representation without per-scene optimization.


\section{Experiments}
\label{sec:exp}

\begin{table*}[t]
\centering
\begingroup
\footnotesize
\setlength{\tabcolsep}{1.7pt}
\begin{tabular*}{\textwidth}{@{}l@{\extracolsep{\fill}}cccccccccccc@{}}
\toprule
\multirow{2}{*}{Method}
 & \multicolumn{4}{c}{$\rho=0.1$}
 & \multicolumn{4}{c}{$\rho=0.01$}
 & \multicolumn{4}{c}{$\rho=0.001$} \\
\cmidrule(lr){2-5}\cmidrule(lr){6-9}\cmidrule(lr){10-13}
 & PSNR\smash{$\uparrow$} & SSIM\smash{$\uparrow$} & LPIPS\smash{$\downarrow$} & Time (s)\smash{$\downarrow$} & PSNR\smash{$\uparrow$} & SSIM\smash{$\uparrow$} & LPIPS\smash{$\downarrow$} & Time (s)\smash{$\downarrow$} & PSNR\smash{$\uparrow$} & SSIM\smash{$\uparrow$} & LPIPS\smash{$\downarrow$} & Time (s)\smash{$\downarrow$} \\
\midrule
\multicolumn{13}{l}{\textit{NeRF Synthetic} \;(3DGS: 33.47 / 0.970 / 0.030)}\\
LightGS & 21.88 & 0.888 & 0.097 & 31.98 & 15.76 & 0.807 & 0.181 & 32.34 & 12.64 & 0.796 & 0.224 & 32.36\\
PUP-3DGS & 20.24 & 0.860 & 0.116 & 29.20 & 13.26 & 0.786 & 0.206 & 29.16 & 11.31 & 0.792 & 0.230 & 28.59\\
GHAP & 21.19 & 0.854 & 0.125 & \underline{6.28} & 13.40 & 0.785 & 0.206 & \underline{6.50} & 11.25 & 0.792 & 0.242 & \underline{3.52}\\
NanoGS & \underline{25.81} & \underline{0.910} & \underline{0.092} & 11.36 & \underline{22.28} & \underline{0.858} & \underline{0.153} & 11.70 & \underline{19.04} & \underline{0.822} & \underline{0.207} & 11.83\\
\textbf{Ours} & \textbf{27.76} & \textbf{0.936} & \textbf{0.079} & \textbf{3.58} & \textbf{23.75} & \textbf{0.880} & \textbf{0.134} & \textbf{3.02} & \textbf{20.18} & \textbf{0.846} & \textbf{0.191} & \textbf{2.75}\\
\midrule
\multicolumn{13}{l}{\textit{Mip-NeRF360} \;(3DGS: 27.43 / 0.813 / 0.221)}\\
LightGS & 19.38 & \underline{0.588} & \underline{0.412} & 76.14  & 14.39 & 0.389 & \underline{0.582} & 74.79 & 11.97 & 0.278 & 0.662 & 74.54\\
PUP-3DGS & 15.96 & 0.537 & 0.428 & 76.36  & 10.89 & 0.232 & 0.623 & 74.73 & 9.32  & 0.100 & 0.691 & 74.22\\
GHAP & 17.35 & 0.444 & 0.494 & \underline{34.23} & 10.62 & 0.174 & 0.664 & \underline{35.24} & 8.52  & 0.035 & 0.728 & \underline{31.40}\\
NanoGS & \underline{21.97} & 0.582 & 0.432 & 149.64 & \underline{19.39} & \underline{0.470} & 0.587 & 158.05 & \underline{17.20} & \underline{0.430} & \underline{0.661} & 158.12\\
\textbf{Ours} & \textbf{23.45} & \textbf{0.665} & \textbf{0.366} & \textbf{13.71} & \textbf{20.57} & \textbf{0.523} & \textbf{0.532} & \textbf{9.01} & \textbf{18.11} & \textbf{0.466} & \textbf{0.626} & \textbf{8.09}\\
\midrule
\multicolumn{13}{l}{\textit{Tanks \& Temples} \;(3DGS: 23.61 / 0.843 / 0.169)}\\
LightGS & 17.50 & \underline{0.642} & \underline{0.357} & 32.15 & 12.30 & 0.452 & 0.585 & 31.91 & 9.36  & 0.341 & 0.690 & 33.30\\
PUP-3DGS & 13.65 & 0.585 & 0.403 & 31.29 & 9.11  & 0.325 & 0.625 & 30.64 & 7.40  & 0.210 & 0.699 & 30.46\\
GHAP & 15.34 & 0.487 & 0.484 & \underline{23.45} & 8.43  & 0.220 & 0.672 & \underline{22.91} & 5.33  & 0.026 & 0.748 & \underline{20.17}\\
NanoGS & \underline{17.94} & 0.626 & 0.413 & 57.84 & \underline{15.29} & \underline{0.501} & \underline{0.576} & 62.03 & \underline{13.54} & \underline{0.457} & \underline{0.641} & 62.33\\
\textbf{Ours} & \textbf{19.66} & \textbf{0.712} & \textbf{0.324} & \textbf{7.53} & \textbf{16.61} & \textbf{0.532} & \textbf{0.539} & \textbf{5.62} & \textbf{14.50} & \textbf{0.471} & \textbf{0.625} & \textbf{5.71}\\
\midrule
\multicolumn{13}{l}{\textit{Deep Blending} \;(3DGS: 29.69 / 0.907 / 0.238)}\\
LightGS & 24.28 & 0.816 & \underline{0.351} & 33.50  & 18.28 & 0.712 & 0.506 & 31.31 & 13.41 & 0.616 & 0.610 & 31.99\\
PUP-3DGS & 19.90 & 0.765 & 0.391 & 33.45 & 10.91 & 0.441 & 0.615 & \underline{31.05} & 8.23  & 0.166 & 0.708 & 30.16\\
GHAP & 21.75 & 0.739 & 0.436 & \underline{31.56} & 11.36 & 0.435 & 0.646 & 31.98 & 7.56  & 0.051 & 0.746 & \underline{29.42}\\
NanoGS & \underline{26.29} & \underline{0.839} & 0.371 & 104.34 & \underline{23.12} & \underline{0.780} & \underline{0.467} & 97.94 & \underline{19.42} & \underline{0.739} & \underline{0.507} & 99.63\\
\textbf{Ours} & \textbf{27.42} & \textbf{0.864} & \textbf{0.317} & \textbf{11.63} & \textbf{24.35} & \textbf{0.808} & \textbf{0.434} & \textbf{7.63} & \textbf{20.52} & \textbf{0.765} & \textbf{0.484} & \textbf{6.97}\\
\bottomrule
\end{tabular*}%
\endgroup
\caption{Quantitative results evaluated on NeRF Synthetic, Mip-NeRF360,
Tanks \& Temples, and Deep Blending datasets. We highlight the best-performing
results in bold and underline the second-best results among all compared methods.}
\label{tab:main}
\end{table*}

\subsection{Experimental Settings}

\paragraph{Evaluation Datasets and Metrics.}
We evaluate on four standard 3DGS benchmarks, including NeRF-Synthetic~\citep{mildenhall2020nerf}, Mip-NeRF360~\citep{barron2022mipnerf360}, Tanks~\&~Temples~\citep{knapitsch2017tanks}, and Deep Blending~\citep{hedman2018deepblending}. They cover synthetic objects, unbounded real scenes, large-scale captures, and indoor scenes. We measure rendering fidelity on test views using PSNR, SSIM~\citep{wang2004image}, and LPIPS~\citep{zhang2018lpips}. We also report end-to-end simplification time.

\paragraph{Baselines and Comparison Protocol.}
We compare our method with LightGS \citep{fan2024lightgaussian},
PUP-3DGS \citep{hanson2024pup3dgs}, GHAP \citep{wang2025ghap},
and NanoGS \citep{xiong2026nanogs}. NanoGS is the closest baseline
because it is also a training-free post-hoc simplification method.
Following NanoGS, we adopt an immediate-output protocol. Each method
is evaluated directly after primitive-count reduction, without
method-specific recovery, refinement, or fine-tuning. Thus, LightGS,
PUP-3DGS, and GHAP are evaluated after their pruning, selection, and
reduction stages, respectively. All methods use the same trained 3DGS
inputs, target count $M=\lceil\rho N\rceil$, renderer/evaluator, and
hardware. We evaluate $\rho\in\{0.1,0.01,0.001\}$.Note that Gaussian compression methods, such as attribute quantization and entropy coding, are orthogonal to our work. Because they compress the parameters of a fixed primitive set rather than reducing the primitive count, these techniques can be seamlessly applied to our simplified output as a subsequent step.

\paragraph{Implementation Details.}
All inputs are standard 3DGS models optimized with the official implementation~\cite{kerbl20233dgs}. MergeNet is trained once for $30$k steps on cells sampled from 3DGS scenes built from LLFF~\citep{mildenhall2019llff}, ShapeNet~\citep{chang2015shapenet}, and ScanNet~\citep{dai2017scannet}. We set $\lambda=0.2$ and
$\lambda_g=0.2$. The same checkpoint is used for all test scenes and simplification ratios. No scene-specific optimization is performed. For CVT construction, we filter out Gaussians with opacity below $\tau_\alpha=0.1$. We then run five Lloyd iterations to optimize Eq.~\eqref{eq:cvt_discrete}. Simplification time covers the full
pipeline, from loading the trained scene to writing the output. All experiments run on a single NVIDIA RTX~4090 GPU
(24\,GB VRAM) with PyTorch~2.8 and CUDA~12.8.


\begin{figure*}[t!]
\centering
\includegraphics[width=\textwidth,  height=17.5cm]{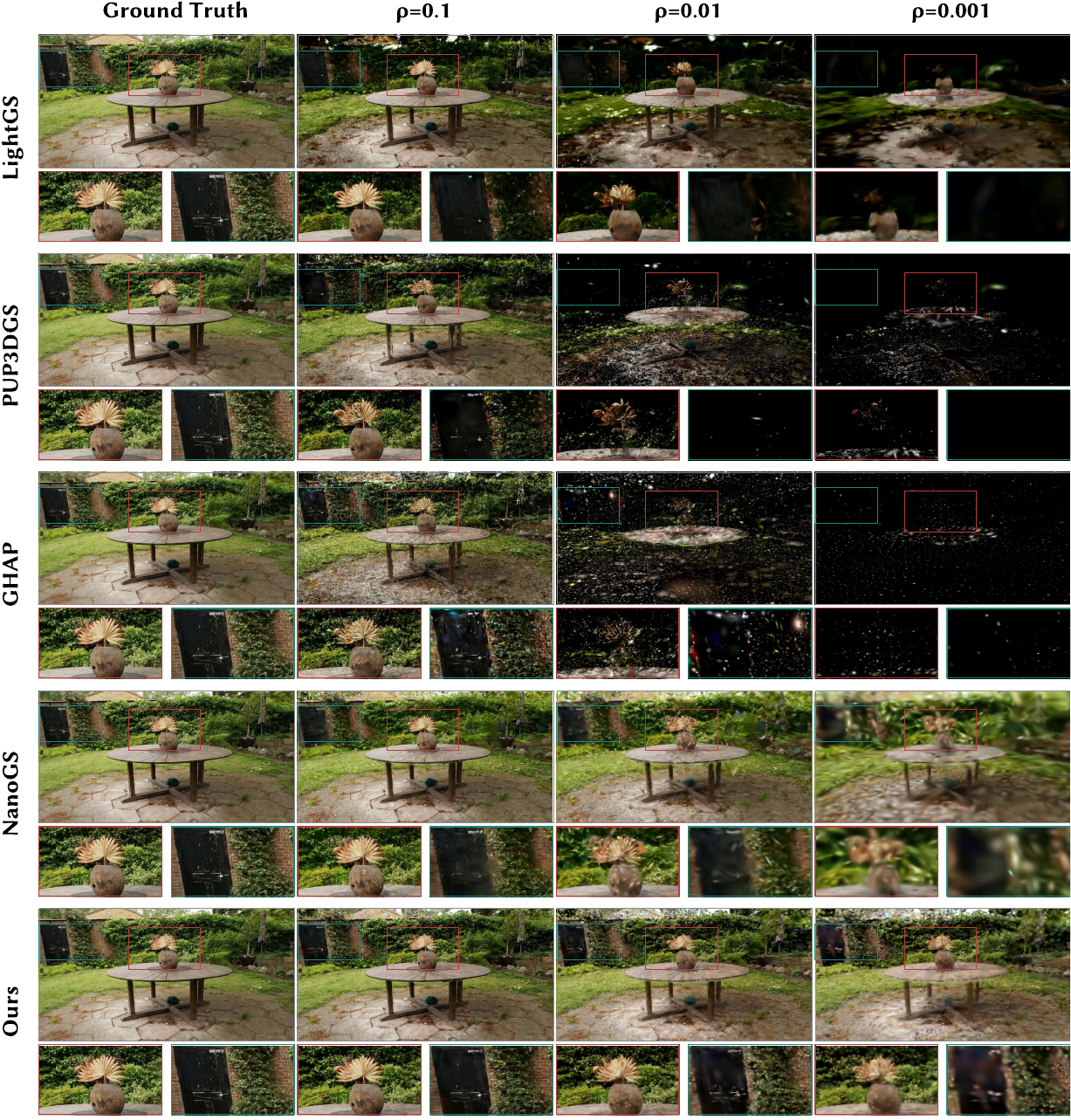}
\caption{Qualitative results on \emph{garden} (Mip-NeRF360).
Comparison of our method with existing baselines under different
simplification ratios.}
\label{fig:qual}
\end{figure*}

\begin{figure*}[t]
\centering
\includegraphics[width=0.7\textwidth,  height=6.5cm]{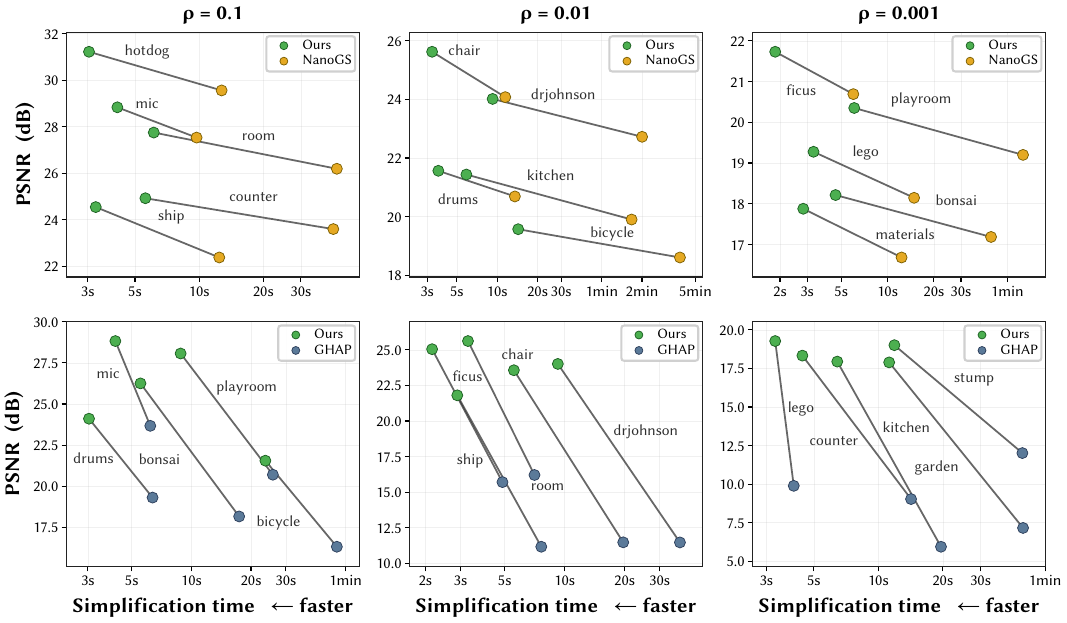}
\caption{Trade-off between quality and simplification time under different simplification ratios.
Top and bottom rows compare our method with NanoGS and GHAP, respectively. Each
connected pair denotes the same scene; upper-left indicates higher PSNR and
shorter end-to-end simplification time.}
\label{fig:pareto}
\end{figure*}

\subsection{Experimental Results}

\paragraph{Quantitative Results.}
Table~\ref{tab:main}
compares our method with LightGS, PUP-3DGS, GHAP, and NanoGS across four benchmarks and three simplification ratios. At matched primitive counts, our method achieves the best rendering quality and the shortest average simplification time in every setting.
Compared with NanoGS, the most relevant SOTA training-free post-hoc simplification baseline, our method improves average PSNR by
$+1.57/+1.30/+1.03$\,dB and SSIM by $+0.055/+0.034/+0.025$ for $\rho=0.1/0.01/0.001$, respectively.

The gains become more significant as the simplification ratio decreases. At $\rho=0.001$, our method outperforms the strongest pruning/selection/reduction baseline among LightGS, PUP-3DGS, and GHAP by $+5.1$ to $+7.5$\,dB PSNR across datasets. On real-captured scenes such as Mip-NeRF360 and Tanks~\&~Temples, pruning and tree-based reduction degrade rapidly at low ratios, whereas our CVT-based support construction maintains more stable fidelity.

\paragraph{Qualitative Results.}
\looseness=-1
Figure~\ref{fig:qual} compares visual quality on the \emph{garden} scene under progressively more aggressive simplification.
The pretrained 3DGS representation of the \emph{garden} scene contains $N=4{,}207{,}352$ Gaussian primitives. Accordingly, the simplification ratios $\rho=0.1$, $0.01$, and $0.001$ yield $420{,}736$, $42{,}074$, and $4{,}208$ output primitives, respectively.
While most methods preserve the coarse layout at $\rho=0.1$, pruning/selection/reduction baselines quickly develop missing regions, speckles, and dark holes at $\rho=0.01$ and $\rho=0.001$.
NanoGS blurs the plant, tabletop, and background structures under aggressive simplification. Our method better preserves the vase silhouette, radial plant details, tabletop contour, and background foliage, showing that CVT support construction and MergeNet retain both global structure and local appearance. Additional qualitative and per-scene results are provided in the supplementary material.

\paragraph{Simplification Time.}
We measure end-to-end simplification time from loading a trained 3DGS scene to writing the simplified output. Figure~\ref{fig:pareto} compares our method with NanoGS and GHAP on matched scenes.
\par\noindent
Using the pooled time over all evaluated scenes and simplification ratios, our method is $12.3\times$ faster than NanoGS and $3.1\times$ faster than GHAP while achieving higher PSNR. These gains come from constructing CVT supports once and predicting the output Gaussians with a single MergeNet pass.
Supplementary comparisons further show $7.2\times$/$7.0\times$ speedups over LightGS/PUP-3DGS.

\subsection{Ablation Studies}
\label{sec:ablation}

In this section, we verify the effectiveness of different components of CVT-GS.
Table~\ref{tab:ablation} ablates CVT support construction, MergeNet, and opacity
filtering using averages over the four benchmarks. All variants use the same
target count $M=\lceil\rho N\rceil$, and the full model performs best at every
simplification ratio. Per-benchmark results are provided in the supplementary material.

\begin{table}[t]
\centering
\setlength{\tabcolsep}{1pt}
\scriptsize
\begin{tabular*}{\columnwidth}{@{}l@{\extracolsep{\fill}}ccccccccc@{}}
\toprule
\multirow{2}{*}{Variant}
& \multicolumn{3}{c}{$\rho=0.1$}
& \multicolumn{3}{c}{$\rho=0.01$}
& \multicolumn{3}{c}{$\rho=0.001$}\\
\cmidrule(lr){2-4}\cmidrule(lr){5-7}\cmidrule(lr){8-10}
& PSNR & SSIM & LPIPS & PSNR & SSIM & LPIPS & PSNR & SSIM & LPIPS\\
\midrule
w/o CVT       & 23.69 & 0.766 & 0.294 & 19.67 & 0.623 & 0.501 & 15.80 & 0.509 & 0.677\\
w/o filtering & 23.73 & 0.772 & 0.294 & 20.56 & 0.660 & 0.441 & 17.47 & 0.610 & 0.530\\
w/o MergeNet  & 22.62 & 0.737 & 0.337 & 18.31 & 0.572 & 0.595 & 14.10 & 0.430 & 0.811\\
\textbf{Full} & \textbf{24.57} & \textbf{0.794} & \textbf{0.272} & \textbf{21.32} & \textbf{0.686} & \textbf{0.410} & \textbf{18.33} & \textbf{0.637} & \textbf{0.481}\\
\bottomrule
\end{tabular*}%
\caption{Quantitative ablation study averaged over four benchmark datasets. We ablate CVT support construction, MergeNet, and opacity filtering. }
\label{tab:ablation}
\end{table}

\paragraph{Effectiveness of CVT Support Construction.}
\looseness=-1
Replacing CVT with local $k$-NN grouping consistently degrades quality, and the
gap widens as the simplification ratio decreases. The average PSNR drops by
$0.88$, $1.65$, and $2.53$\,dB at $\rho=0.1$, $0.01$, and $0.001$,
respectively, while average LPIPS increases from $0.481$ to $0.677$ at the most
aggressive ratio. This trend supports the central premise of our method. When
one output Gaussian must summarize many input primitives, the support should be
formed by a scene-level density-adaptive partition rather than by independent
local groups.

\paragraph{Effectiveness of MergeNet and Filtering.}
Removing MergeNet causes the largest degradation in most settings, with average
PSNR drops of $1.95$, $3.01$, and $4.23$\,dB across the three ratios. This
indicates that the learned neural cell merger is the key component for
high-fidelity many-to-one Gaussian merging. Deterministic cell statistics provide
a stable reference but cannot capture the residual geometry and appearance
within each cell. Opacity filtering has a smaller but consistent effect,
improving average PSNR by $0.76$--$0.86$\,dB and reducing LPIPS by
$0.022$--$0.049$.

\section{Conclusion}

In this paper, we presented \method{}, an optimization-free, post-hoc simplification framework for pre-trained 3DGS scenes. By reformulating simplification as scene-level CVT support allocation followed by rendering-aware many-to-one merging, \method{} moves beyond naive primitive removal. It delivers superior visual quality while operating more than one order of magnitude faster than existing baselines, all while maintaining strict compatibility with the standard 3DGS format.
Specifically, our geometry-aware CVT allocates coherent, density-adaptive support cells over Gaussian centers, which MergeNet then condenses into single representative primitives in a single feed-forward pass. 
Extensive experiments across four standard benchmarks demonstrate that \method{} substantially improves the trade-off between synthesis quality and simplification speed under aggressive compression ratios. Notably, at the extreme $100\times$ simplification setting, \method{} achieves an average PSNR of 21.32 dB, outperforming the sota post-hoc baseline by 1.30 dB. 
Ablation studies further confirm the indispensability of both CVT spatial partitioning and neural cell merging.
For future work, researchers may explore more sophisticated MergeNet architectures and extend this framework to dynamic 4D representations where temporal dimensions are integrated.
\clearpage
\bibliography{cvtgs}

\begin{thebibliography}{34}
\providecommand{\natexlab}[1]{#1}

\bibitem[{Barron et~al.(2021)Barron, Mildenhall, Tancik, Hedman,
  Martin-Brualla, and Srinivasan}]{barron2021mipnerf}
Barron, J.~T.; Mildenhall, B.; Tancik, M.; Hedman, P.; Martin-Brualla, R.; and
  Srinivasan, P.~P. 2021.
\newblock Mip-{N}e{RF}: A Multiscale Representation for Anti-Aliasing Neural
  Radiance Fields.
\newblock In \emph{Proceedings of the IEEE/CVF International Conference on
  Computer Vision (ICCV)}.

\bibitem[{Barron et~al.(2022)Barron, Mildenhall, Verbin, Srinivasan, and
  Hedman}]{barron2022mipnerf360}
Barron, J.~T.; Mildenhall, B.; Verbin, D.; Srinivasan, P.~P.; and Hedman, P.
  2022.
\newblock Mip-{N}e{RF} 360: Unbounded Anti-Aliased Neural Radiance Fields.
\newblock In \emph{Proceedings of the IEEE/CVF Conference on Computer Vision
  and Pattern Recognition (CVPR)}.

\bibitem[{Chang et~al.(2015)Chang, Funkhouser, Guibas, Hanrahan, Huang, Li,
  Savarese, Savva, Song, Su, Xiao, Yi, and Yu}]{chang2015shapenet}
Chang, A.~X.; Funkhouser, T.; Guibas, L.; Hanrahan, P.; Huang, Q.; Li, Z.;
  Savarese, S.; Savva, M.; Song, S.; Su, H.; Xiao, J.; Yi, L.; and Yu, F. 2015.
\newblock {ShapeNet}: An Information-Rich 3{D} Model Repository.
\newblock Technical Report arXiv:1512.03012, Stanford University, Princeton
  University, Toyota Technological Institute at Chicago.

\bibitem[{Chen et~al.(2026)Chen, Li, Wu, Lin, Harandi, and Cai}]{chen2026pcgs}
Chen, Y.; Li, M.; Wu, Q.; Lin, W.; Harandi, M.; and Cai, J. 2026.
\newblock {PCGS}: Progressive Compression of 3{D} {G}aussian Splatting.
\newblock In \emph{Proceedings of the AAAI Conference on Artificial
  Intelligence (AAAI)}.

\bibitem[{Chen et~al.(2025)Chen, Wu, Li, Lin, Harandi, and
  Cai}]{chen2025fastfeedforward}
Chen, Y.; Wu, Q.; Li, M.; Lin, W.; Harandi, M.; and Cai, J. 2025.
\newblock Fast Feedforward 3{D} {G}aussian Splatting Compression.
\newblock In \emph{International Conference on Learning Representations
  (ICLR)}.

\bibitem[{Chen et~al.(2024)Chen, Wu, Lin, Harandi, and Cai}]{chen2024hac}
Chen, Y.; Wu, Q.; Lin, W.; Harandi, M.; and Cai, J. 2024.
\newblock {HAC}: Hash-grid Assisted Context for 3{D} {G}aussian Splatting
  Compression.
\newblock In \emph{European Conference on Computer Vision (ECCV)}.

\bibitem[{Dai et~al.(2017)Dai, Chang, Savva, Halber, Funkhouser, and
  Nie{\ss}ner}]{dai2017scannet}
Dai, A.; Chang, A.~X.; Savva, M.; Halber, M.; Funkhouser, T.; and Nie{\ss}ner,
  M. 2017.
\newblock {ScanNet}: Richly-Annotated 3{D} Reconstructions of Indoor Scenes.
\newblock In \emph{Proceedings of the IEEE Conference on Computer Vision and
  Pattern Recognition (CVPR)}.

\bibitem[{Dai, Liu, and Zhang(2025)}]{dai2025decoupled}
Dai, Z.; Liu, T.; and Zhang, Y. 2025.
\newblock Efficient Decoupled Feature 3{D} {G}aussian Splatting via
  Hierarchical Compression.
\newblock In \emph{Proceedings of the IEEE/CVF Conference on Computer Vision
  and Pattern Recognition (CVPR)}.

\bibitem[{Du, Faber, and Gunzburger(1999)}]{du1999cvt}
Du, Q.; Faber, V.; and Gunzburger, M. 1999.
\newblock Centroidal {V}oronoi Tessellations: Applications and Algorithms.
\newblock \emph{SIAM Review}, 41(4).

\bibitem[{Fan et~al.(2024)Fan, Wang, Wen, Zhu, Xu, and
  Wang}]{fan2024lightgaussian}
Fan, Z.; Wang, K.; Wen, K.; Zhu, Z.; Xu, D.; and Wang, Z. 2024.
\newblock {L}ight{G}aussian: Unbounded 3{D} {G}aussian Compression with 15x
  Reduction and 200+ {FPS}.
\newblock In \emph{Advances in Neural Information Processing Systems
  (NeurIPS)}.

\bibitem[{Fang and Wang(2024)}]{fang2024minisplatting}
Fang, G.; and Wang, B. 2024.
\newblock Mini-Splatting: Representing Scenes with a Constrained Number of
  Gaussians.
\newblock In \emph{European Conference on Computer Vision (ECCV)}.

\bibitem[{Girish, Gupta, and Shrivastava(2024)}]{girish2024eagles}
Girish, S.; Gupta, K.; and Shrivastava, A. 2024.
\newblock {EAGLES}: Efficient Accelerated 3{D} {G}aussians with Lightweight
  Encodings.
\newblock In \emph{European Conference on Computer Vision (ECCV)}.

\bibitem[{Hanson et~al.(2025)Hanson, Tu, Singla, Jayawardhana, Zwicker, and
  Goldstein}]{hanson2024pup3dgs}
Hanson, A.; Tu, A.; Singla, V.; Jayawardhana, M.; Zwicker, M.; and Goldstein,
  T. 2025.
\newblock {PUP} 3{D}-{GS}: Principled Uncertainty Pruning for 3{D} {G}aussian
  Splatting.
\newblock In \emph{Proceedings of the IEEE/CVF Conference on Computer Vision
  and Pattern Recognition (CVPR)}.

\bibitem[{Hedman et~al.(2018)Hedman, Philip, Price, Frahm, Drettakis, and
  Brostow}]{hedman2018deepblending}
Hedman, P.; Philip, J.; Price, T.; Frahm, J.-M.; Drettakis, G.; and Brostow, G.
  2018.
\newblock Deep Blending for Free-Viewpoint Image-Based Rendering.
\newblock \emph{ACM Transactions on Graphics}, 37(6).

\bibitem[{Kerbl et~al.(2023)Kerbl, Kopanas, Leimk{\"u}hler, and
  Drettakis}]{kerbl20233dgs}
Kerbl, B.; Kopanas, G.; Leimk{\"u}hler, T.; and Drettakis, G. 2023.
\newblock 3{D} {G}aussian Splatting for Real-Time Radiance Field Rendering.
\newblock In \emph{ACM Transactions on Graphics (SIGGRAPH)}, volume~42.

\bibitem[{Knapitsch et~al.(2017)Knapitsch, Park, Zhou, and
  Koltun}]{knapitsch2017tanks}
Knapitsch, A.; Park, J.; Zhou, Q.-Y.; and Koltun, V. 2017.
\newblock Tanks and Temples: Benchmarking Large-Scale Scene Reconstruction.
\newblock \emph{ACM Transactions on Graphics}, 36(4).

\bibitem[{Lee et~al.(2024)Lee, Rho, Sun, Ko, and Park}]{lee2024compact3dgs}
Lee, J.~C.; Rho, D.; Sun, X.; Ko, J.~H.; and Park, E. 2024.
\newblock Compact 3{D} {G}aussian Representation for Radiance Field.
\newblock In \emph{Proceedings of the IEEE/CVF Conference on Computer Vision
  and Pattern Recognition (CVPR)}.

\bibitem[{L{\'e}vy(2026)}]{geogram}
L{\'e}vy, B. 2026.
\newblock {Geogram}: A Programming Library with Geometric Algorithms.
\newblock \url{https://github.com/BrunoLevy/geogram}.
\newblock Accessed: 2026-07-05.

\bibitem[{L{\'e}vy and Liu(2010)}]{levy2010lpcvt}
L{\'e}vy, B.; and Liu, Y. 2010.
\newblock {Lp} Centroidal {V}oronoi Tessellation and Its Applications.
\newblock \emph{ACM Transactions on Graphics (SIGGRAPH)}, 29(4).

\bibitem[{Liu et~al.(2025)Liu, Wang, Li, Cai, Wang, Li, Molchanov, Wang, and
  Wang}]{liu2025flexgs}
Liu, H.; Wang, Y.; Li, C.; Cai, R.; Wang, K.; Li, W.; Molchanov, P.; Wang, P.;
  and Wang, Z. 2025.
\newblock {FlexGS}: Train Once, Deploy Everywhere with Many-in-One Flexible
  3{D} {G}aussian Splatting.
\newblock In \emph{Proceedings of the IEEE/CVF Conference on Computer Vision
  and Pattern Recognition (CVPR)}.

\bibitem[{Liu et~al.(2009)Liu, Wang, L{\'e}vy, Sun, Yan, Lu, and
  Yang}]{liu2009cvt}
Liu, Y.; Wang, W.; L{\'e}vy, B.; Sun, F.; Yan, D.-M.; Lu, L.; and Yang, C.
  2009.
\newblock On Centroidal {V}oronoi Tessellation---Energy Smoothness and Fast
  Computation.
\newblock \emph{ACM Transactions on Graphics}, 28(4).

\bibitem[{Lloyd(1982)}]{lloyd1982least}
Lloyd, S. 1982.
\newblock Least Squares Quantization in {PCM}.
\newblock \emph{IEEE Transactions on Information Theory}, 28(2).

\bibitem[{Lu et~al.(2024)Lu, Yu, Xu, Xiangli, Wang, Lin, and
  Dai}]{lu2024scaffold}
Lu, T.; Yu, M.; Xu, L.; Xiangli, Y.; Wang, L.; Lin, D.; and Dai, B. 2024.
\newblock Scaffold-{GS}: Structured 3{D} {G}aussians for View-Adaptive
  Rendering.
\newblock In \emph{Proceedings of the IEEE/CVF Conference on Computer Vision
  and Pattern Recognition (CVPR)}.

\bibitem[{Mildenhall et~al.(2019)Mildenhall, Srinivasan, Ortiz-Cayon,
  Kalantari, Ramamoorthi, Ng, and Kar}]{mildenhall2019llff}
Mildenhall, B.; Srinivasan, P.~P.; Ortiz-Cayon, R.; Kalantari, N.~K.;
  Ramamoorthi, R.; Ng, R.; and Kar, A. 2019.
\newblock Local Light Field Fusion: Practical View Synthesis with Prescriptive
  Sampling Guidelines.
\newblock \emph{ACM Transactions on Graphics}, 38(4).

\bibitem[{Mildenhall et~al.(2020)Mildenhall, Srinivasan, Tancik, Barron,
  Ramamoorthi, and Ng}]{mildenhall2020nerf}
Mildenhall, B.; Srinivasan, P.~P.; Tancik, M.; Barron, J.~T.; Ramamoorthi, R.;
  and Ng, R. 2020.
\newblock Ne{RF}: Representing Scenes as Neural Radiance Fields for View
  Synthesis.
\newblock In \emph{European Conference on Computer Vision (ECCV)}.

\bibitem[{Navaneet et~al.(2024)Navaneet, Pourahmadi~Meibodi,
  Abbasi~Koohpayegani, and Pirsiavash}]{navaneet2024compgs}
Navaneet, K.~L.; Pourahmadi~Meibodi, K.; Abbasi~Koohpayegani, S.; and
  Pirsiavash, H. 2024.
\newblock Comp{GS}: Smaller and Faster {G}aussian Splatting with Vector
  Quantization.
\newblock In \emph{European Conference on Computer Vision (ECCV)}.

\bibitem[{Niedermayr, Stumpfegger, and
  Westermann(2024)}]{niedermayr2024compressed3dgs}
Niedermayr, S.; Stumpfegger, J.; and Westermann, R. 2024.
\newblock Compressed 3{D} {G}aussian Splatting for Accelerated Novel View
  Synthesis.
\newblock In \emph{Proceedings of the IEEE/CVF Conference on Computer Vision
  and Pattern Recognition (CVPR)}.

\bibitem[{Tang et~al.(2026)Tang, Feng, Cheng, Yu, Zhang, Liu, Long, Wang, and
  Yuan}]{tang2026neuralgs}
Tang, Z.; Feng, C.; Cheng, X.; Yu, W.; Zhang, J.; Liu, Y.; Long, X.-X.; Wang,
  W.; and Yuan, L. 2026.
\newblock {NeuralGS}: Bridging Neural Fields and 3{D} {G}aussian Splatting for
  Compact 3{D} Representations.
\newblock In \emph{Proceedings of the AAAI Conference on Artificial
  Intelligence (AAAI)}.

\bibitem[{Wang et~al.(2025)Wang, Li, Zeng, Meng, and Zhang}]{wang2025ghap}
Wang, T.; Li, M.; Zeng, G.; Meng, C.; and Zhang, Q. 2025.
\newblock {G}aussian Herding across Pens: An Optimal Transport Perspective on
  Global {G}aussian Reduction for 3{DGS}.
\newblock NeurIPS 2025, arXiv:2506.09534.

\bibitem[{Wang et~al.(2004)Wang, Bovik, Sheikh, and Simoncelli}]{wang2004image}
Wang, Z.; Bovik, A.~C.; Sheikh, H.~R.; and Simoncelli, E.~P. 2004.
\newblock Image Quality Assessment: From Error Visibility to Structural
  Similarity.
\newblock \emph{IEEE Transactions on Image Processing}, 13(4): 600--612.

\bibitem[{Xiong et~al.(2026)Xiong, Liu, Zhou, Chen, Fan, and
  Feng}]{xiong2026nanogs}
Xiong, B.; Liu, R.; Zhou, T.; Chen, M.; Fan, Z.; and Feng, A. 2026.
\newblock Nano{GS}: Training-Free {G}aussian Splat Simplification.
\newblock arXiv:2603.16103.

\bibitem[{Zhan et~al.(2025)Zhan, Ho, Yang, Chen, Chiang, Liu, and
  Peng}]{zhan2025cat3dgs}
Zhan, Y.-T.; Ho, C.-Y.; Yang, H.; Chen, Y.-H.; Chiang, J.~C.; Liu, Y.-L.; and
  Peng, W.-H. 2025.
\newblock {CAT}-3{DGS}: A Context-Adaptive Triplane Approach to
  Rate-Distortion-Optimized 3{DGS} Compression.
\newblock In \emph{International Conference on Learning Representations
  (ICLR)}.

\bibitem[{Zhang et~al.(2018)Zhang, Isola, Efros, Shechtman, and
  Wang}]{zhang2018lpips}
Zhang, R.; Isola, P.; Efros, A.~A.; Shechtman, E.; and Wang, O. 2018.
\newblock The Unreasonable Effectiveness of Deep Features as a Perceptual
  Metric.
\newblock In \emph{Proceedings of the IEEE Conference on Computer Vision and
  Pattern Recognition (CVPR)}.

\bibitem[{Zhang et~al.(2024)Zhang, Song, Lee, Yang, Peng, Chellappa, and
  Fan}]{zhang2024lp3dgs}
Zhang, Z.; Song, T.; Lee, Y.; Yang, L.; Peng, C.; Chellappa, R.; and Fan, D.
  2024.
\newblock {LP}-3{DGS}: Learning to Prune 3{D} {G}aussian Splatting.
\newblock In \emph{Advances in Neural Information Processing Systems
  (NeurIPS)}.

\end{thebibliography}

\clearpage
\onecolumn
\section*{Supplementary Material for CVT-GS}

\begin{center}
\centering
\includegraphics[width=\textwidth,height=1.10\textwidth,keepaspectratio]{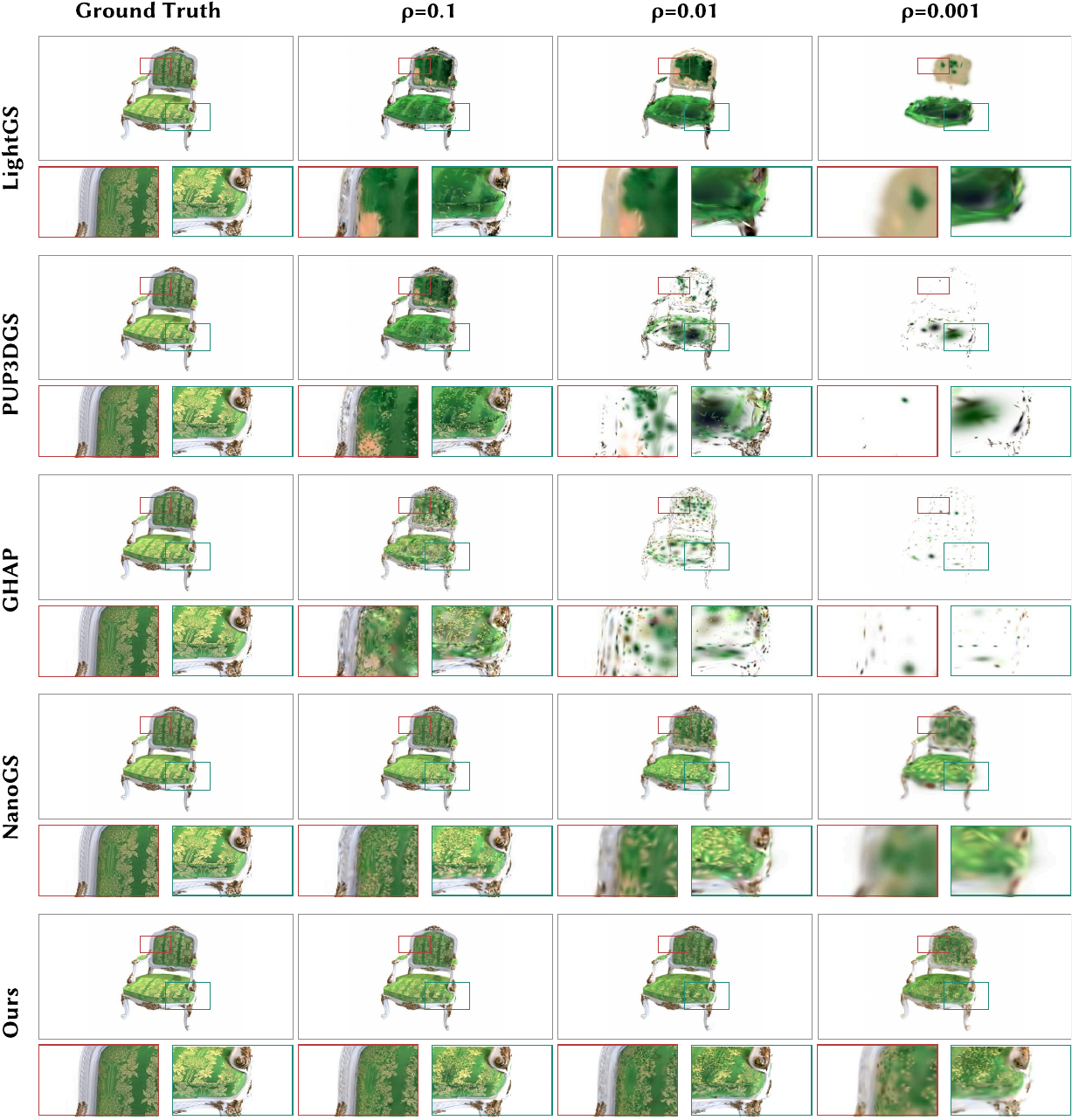}
\captionof{figure}{Qualitative results on \emph{chair} (NeRF-Synthetic).
Comparison of our method with existing baselines under different
simplification ratios.}
\label{fig:qual_chair}
\end{center}

\clearpage

\begin{center}
\centering
\includegraphics[width=0.64\textwidth]{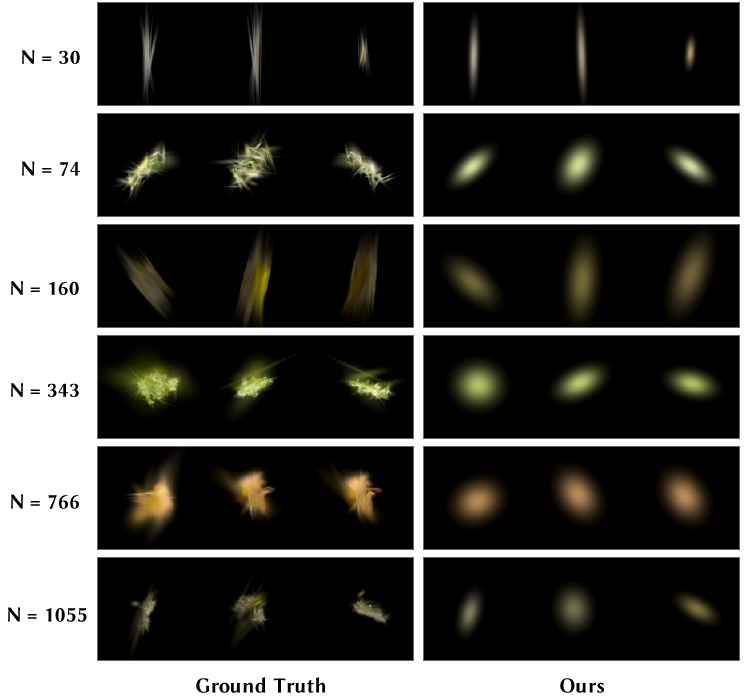}
\captionof{figure}{Cell-level rendering examples of the learned many-to-one
merger. Each row shows one CVT support cell containing a variable number of input Gaussian
primitives; the left column renders the original cell, and the right column
renders the one-Gaussian output of our method.}
\label{fig:cell_splat}
\end{center}

\begin{center}
\centering
\includegraphics[width=0.82\textwidth]{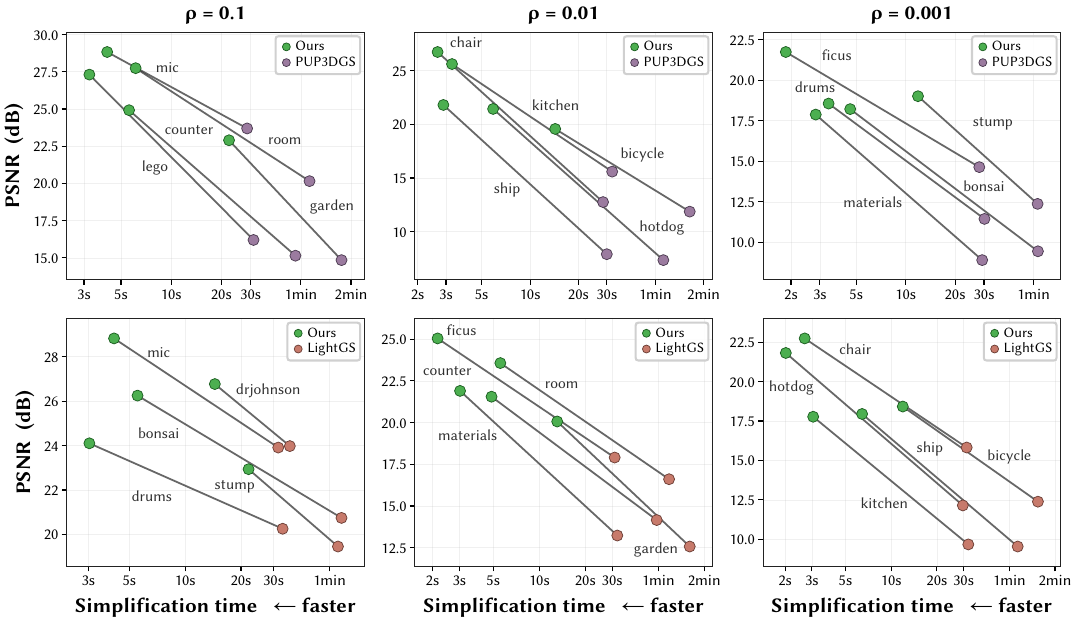}
\captionof{figure}{Trade-off between quality and simplification time under different
simplification ratios. Top and bottom rows compare our method with PUP-3DGS and
LightGS, respectively. Each connected pair denotes the same scene; upper-left
indicates higher PSNR and shorter end-to-end simplification time.}
\label{fig:supp_timing_light_pup}
\end{center}

\clearpage

{\centering
\begingroup
\footnotesize
\renewcommand{\arraystretch}{1.12}
\setlength{\tabcolsep}{4.0pt}
\setlength{\aboverulesep}{0.18ex}
\setlength{\belowrulesep}{0.18ex}
\setlength{\cmidrulekern}{0.25em}
\begin{tabular*}{\linewidth}{@{}l@{\hspace{0.8em}}c@{\extracolsep{\fill}}cc ccc ccc@{}}
\toprule
\multirow{2}{*}{Variant}
 & \multicolumn{3}{c}{$\rho=0.1$}
 & \multicolumn{3}{c}{$\rho=0.01$}
 & \multicolumn{3}{c}{$\rho=0.001$} \\
\cmidrule(lr){2-4}\cmidrule(lr){5-7}\cmidrule(lr){8-10}
 & PSNR$\uparrow$ & SSIM$\uparrow$ & LPIPS$\downarrow$
 & PSNR$\uparrow$ & SSIM$\uparrow$ & LPIPS$\downarrow$
 & PSNR$\uparrow$ & SSIM$\uparrow$ & LPIPS$\downarrow$ \\
\midrule
\addlinespace[0.45ex]
\multicolumn{10}{@{}l}{\textit{NeRF Synthetic}}\\
\addlinespace[0.30ex]
w/o CVT        & 27.12 & 0.926 & 0.088 & 22.45 & 0.845 & 0.183 & 18.35 & 0.792 & 0.275\\
w/o filtering  & 27.01 & 0.929 & 0.085 & 23.71 & 0.873 & 0.143 & 19.81 & 0.840 & 0.198\\
w/o MergeNet   & 26.12 & 0.914 & 0.103 & 20.62 & 0.805 & 0.242 & 15.67 & 0.684 & 0.441\\
\textbf{Full}  & \textbf{27.76} & \textbf{0.936} & \textbf{0.079} & \textbf{23.75} & \textbf{0.880} & \textbf{0.134} & \textbf{20.18} & \textbf{0.846} & \textbf{0.191}\\
\midrule
\addlinespace[0.45ex]
\multicolumn{10}{@{}l}{\textit{Mip-NeRF360}}\\
\addlinespace[0.30ex]
w/o CVT        & 22.73 & 0.631 & 0.384 & 19.12 & 0.473 & 0.595 & 15.63 & 0.374 & 0.779\\
w/o filtering  & 21.86 & 0.615 & 0.424 & 19.08 & 0.471 & 0.599 & 16.74 & 0.419 & 0.686\\
w/o MergeNet   & 21.47 & 0.603 & 0.442 & 18.32 & 0.439 & 0.654 & 14.96 & 0.352 & 0.815\\
\textbf{Full}  & \textbf{23.45} & \textbf{0.665} & \textbf{0.366} & \textbf{20.57} & \textbf{0.523} & \textbf{0.532} & \textbf{18.11} & \textbf{0.466} & \textbf{0.626}\\
\midrule
\addlinespace[0.45ex]
\multicolumn{10}{@{}l}{\textit{Tanks \& Temples}}\\
\addlinespace[0.30ex]
w/o CVT        & 18.72 & 0.679 & 0.357 & 14.97 & 0.462 & 0.648 & 11.43 & 0.282 & 0.935\\
w/o filtering  & 19.20 & 0.697 & 0.338 & 15.87 & 0.514 & 0.562 & 13.61 & 0.438 & 0.714\\
w/o MergeNet   & 17.95 & 0.644 & 0.401 & 13.78 & 0.402 & 0.792 &  9.82 & 0.198 & 1.127\\
\textbf{Full}  & \textbf{19.66} & \textbf{0.712} & \textbf{0.324} & \textbf{16.61} & \textbf{0.532} & \textbf{0.539} & \textbf{14.50} & \textbf{0.471} & \textbf{0.625}\\
\midrule
\addlinespace[0.45ex]
\multicolumn{10}{@{}l}{\textit{Deep Blending}}\\
\addlinespace[0.30ex]
w/o CVT        & 26.18 & 0.827 & 0.349 & 22.15 & 0.710 & 0.578 & 17.81 & 0.588 & 0.719\\
w/o filtering  & 26.85 & 0.848 & 0.330 & 23.57 & 0.784 & 0.458 & 19.74 & 0.742 & 0.523\\
w/o MergeNet   & 24.93 & 0.785 & 0.403 & 20.52 & 0.643 & 0.693 & 15.94 & 0.488 & 0.862\\
\textbf{Full}  & \textbf{27.42} & \textbf{0.864} & \textbf{0.317} & \textbf{24.35} & \textbf{0.808} & \textbf{0.434} & \textbf{20.52} & \textbf{0.765} & \textbf{0.484}\\
\bottomrule
\end{tabular*}
\endgroup
\begingroup
\setlength{\abovecaptionskip}{2pt}
\setlength{\belowcaptionskip}{0pt}
\captionof{table}{Quantitative ablation study on each benchmark
dataset. We ablate CVT support construction, MergeNet, and opacity filtering.
Best results are highlighted in bold.}
\label{tab:supp_ablation}
\endgroup

\par\bigskip
\centering
\begingroup
\footnotesize
\renewcommand{\arraystretch}{1.12}
\setlength{\tabcolsep}{4.0pt}
\setlength{\aboverulesep}{0.18ex}
\setlength{\belowrulesep}{0.18ex}
\setlength{\cmidrulekern}{0.25em}
\begin{tabular*}{\textwidth}{@{\hspace{0.035\textwidth}}l@{\hspace{1.0em}}c@{\extracolsep{\fill}}cc ccc ccc@{\hspace{0.035\textwidth}}}
\toprule
\multirow{2}{*}{Scene}
 & \multicolumn{3}{c}{$\rho=0.1$}
 & \multicolumn{3}{c}{$\rho=0.01$}
 & \multicolumn{3}{c}{$\rho=0.001$} \\
\cmidrule(lr){2-4}\cmidrule(lr){5-7}\cmidrule(lr){8-10}
 & PSNR$\uparrow$ & SSIM$\uparrow$ & LPIPS$\downarrow$
 & PSNR$\uparrow$ & SSIM$\uparrow$ & LPIPS$\downarrow$
 & PSNR$\uparrow$ & SSIM$\uparrow$ & LPIPS$\downarrow$ \\
\midrule
\addlinespace[0.45ex]
\multicolumn{10}{@{\hspace{0.035\textwidth}}l}{\textit{NeRF Synthetic}}\\
\addlinespace[0.30ex]
drums     & 24.1045 & 0.9239 & 0.0689 & 21.5618 & 0.8725 & 0.1291 & 18.5562 & 0.8405 & 0.1990\\
lego      & 27.3102 & 0.9199 & 0.0917 & 22.0116 & 0.8271 & 0.1712 & 19.2768 & 0.7978 & 0.2193\\
materials & 26.2606 & 0.9265 & 0.0919 & 21.8998 & 0.8614 & 0.1430 & 17.8781 & 0.8112 & 0.2192\\
mic       & 28.8284 & 0.9846 & 0.0287 & 25.2896 & 0.9462 & 0.0665 & 21.6724 & 0.9227 & 0.0987\\
ship      & 24.5378 & 0.8202 & 0.1852 & 21.8053 & 0.7711 & 0.2764 & 17.7706 & 0.7428 & 0.3389\\
chair     & 28.7003 & 0.9417 & 0.0544 & 25.6197 & 0.9058 & 0.1013 & 22.7328 & 0.8691 & 0.1591\\
ficus     & 31.1139 & 0.9930 & 0.0293 & 25.0480 & 0.9347 & 0.0617 & 21.7319 & 0.8998 & 0.1164\\
hotdog    & 31.2173 & 0.9738 & 0.0801 & 26.7425 & 0.9246 & 0.1191 & 21.8104 & 0.8830 & 0.1744\\
\midrule
\addlinespace[0.45ex]
\multicolumn{10}{@{\hspace{0.035\textwidth}}l}{\textit{Mip-NeRF360}}\\
\addlinespace[0.30ex]
bicycle  & 21.5512 & 0.5594 & 0.4016 & 19.5702 & 0.3924 & 0.5693 & 18.4167 & 0.3665 & 0.6929\\
flowers  & 18.9185 & 0.4407 & 0.4709 & 17.2129 & 0.3099 & 0.6279 & 15.6628 & 0.2659 & 0.7375\\
garden   & 22.8975 & 0.6338 & 0.3283 & 20.0644 & 0.4055 & 0.5541 & 17.8939 & 0.3442 & 0.7180\\
stump    & 22.9398 & 0.5984 & 0.3755 & 20.4165 & 0.4293 & 0.5701 & 19.0091 & 0.3947 & 0.6832\\
treehill & 20.3475 & 0.4889 & 0.4834 & 19.4054 & 0.4189 & 0.6112 & 18.1206 & 0.3993 & 0.6426\\
room     & 27.7428 & 0.8313 & 0.3176 & 23.5693 & 0.7632 & 0.4348 & 19.3816 & 0.7057 & 0.5032\\
counter  & 24.9181 & 0.8095 & 0.3131 & 21.5521 & 0.6847 & 0.4641 & 18.3329 & 0.6048 & 0.5244\\
kitchen  & 25.4750 & 0.7949 & 0.3096 & 21.4345 & 0.5997 & 0.4848 & 17.9453 & 0.5051 & 0.6069\\
bonsai   & 26.2531 & 0.8305 & 0.2954 & 21.8816 & 0.7051 & 0.4726 & 18.2148 & 0.6118 & 0.5225\\
\midrule
\addlinespace[0.45ex]
\multicolumn{10}{@{\hspace{0.035\textwidth}}l}{\textit{Tanks \& Temples}}\\
\addlinespace[0.30ex]
truck & 21.2978 & 0.7672 & 0.2921 & 17.9508 & 0.5705 & 0.5133 & 15.3810 & 0.4975 & 0.6359\\
train & 18.0144 & 0.6577 & 0.3567 & 15.2701 & 0.4930 & 0.5647 & 13.6154 & 0.4439 & 0.6143\\
\midrule
\addlinespace[0.45ex]
\multicolumn{10}{@{\hspace{0.035\textwidth}}l}{\textit{Deep Blending}}\\
\addlinespace[0.30ex]
drjohnson & 26.7728 & 0.8577 & 0.3196 & 24.0122 & 0.7978 & 0.4484 & 20.6855 & 0.7520 & 0.4894\\
playroom  & 28.0718 & 0.8699 & 0.3138 & 24.6901 & 0.8183 & 0.4188 & 20.3519 & 0.7778 & 0.4782\\
\bottomrule
\end{tabular*}
\endgroup
\begingroup
\setlength{\abovecaptionskip}{2pt}
\setlength{\belowcaptionskip}{0pt}
\captionof{table}{Per-scene quantitative results of our method across four
benchmark datasets and three simplification ratios. Higher PSNR/SSIM and lower
LPIPS indicate better performance.}
\label{tab:perscene}
\endgroup
\par}

\end{document}